\documentclass[twoside,11pt]{article}

\usepackage{jmlr2e}
\usepackage{subfig}
\usepackage{xspace}
\usepackage{paralist}
\usepackage{booktabs}

\newcommand{\lc}{learning curves\xspace}
\newcommand{\meta}{meta-features\xspace}
\newcommand{\mf}{multi-fidelity\xspace}
\newcommand{\as}{Algorithm Selection\xspace}

\newcommand{\hpo}{Hyperparameter Optimization\xspace}
\newcommand{\metal}{meta-learning\xspace}
\newcommand{\mfo}{multi-fidelity optimization\xspace}
\newcommand{\ground}{ground-truth\xspace}
\newcommand{\sh}{Successive Halving\xspace}
\newcommand{\imfas}{\texttt{IMFAS}\xspace}

\jmlrheading{1}{2021}{1-48}{4/00}{10/00}{Mohan, Ruhkopf and Lindauer}

\ShortHeadings{IMFAS}{Mohan, Ruhkopf and Lindauer}
\firstpageno{1}

\begin{document}

\title{Towards Meta-learned Algorithm Selection\\ using Implicit Fidelity Information}

\author{\name Aditya Mohan \thanks{These authors equally contributed to the work} \email mohan@tnt.uni-hannover.de \\
        \name Tim Ruhkopf $^*$ \email ruhkopf@tnt.uni-hannover.de\\
        \name Marius Lindauer \email lindauer@tnt.uni-hannover.de\\
        \addr Leibniz University Hannover
}

\editor{TBA}

\maketitle

\begin{abstract}

  Automatically selecting the best performing algorithm for a given dataset or ranking multiple algorithms by their expected performance supports users in developing new machine learning applications. Most approaches for this problem rely on pre-computed dataset \meta and landmarking performances to capture the salient topology of the datasets and those topologies that the algorithms attend to. Landmarking usually exploits cheap algorithms not necessarily in the pool of candidate algorithms to get inexpensive approximations of the topology. While somewhat indicative, hand-crafted dataset \meta and landmarks are likely insufficient descriptors, strongly depending on the alignment of the geometries that the landmarks and the candidate algorithms search for. We propose \imfas, a method to exploit \mf landmarking information directly from the candidate algorithms in the form of non-parametrically non-myopic meta-learned learning curves via LSTMs in a few-shot setting during testing. Using this mechanism, \imfas jointly learns a dataset’s topology and algorithms’ inductive biases, without the need to expensively train them to convergence. Our approach produces informative landmarks, easily enriched by arbitrary meta-features at a low computational cost, capable of producing the desired ranking using cheaper fidelities. We additionally show that \imfas is able to beat \sh with at most 50\% of the fidelity sequence during test time.
  
\end{abstract}

\section{Introduction}
Selecting well-performing algorithms for a given task in an automatic and data-driven manner, i.e. the \as problem \citep{rice76a}, is considered one of the major challenges in AutoML. The solution to this problem promises to alleviate the necessity to re-evaluate candidate algorithms on every new task/dataset, to figure out either a single best algorithm \citep{leite-ecai10a} or recommend multiple algorithms ranked by their likely performances \citep{rijn-ida15, mohr-arxiv21}. The standard attempt at solving this problem is to create a meta-dataset containing the datasets and the corresponding candidate algorithm performances. A recommender can, then, learn to predict what co-occurrences likely perform well. The performance of this recommender system is tied to how well the topology of the dataset is encoded into its feature space and how indicative these features are for the performance of an algorithm on this dataset. This encoding of the topology is computed through functions/estimators on the dataset such as counting the number of observations, the number of features, computing the class imbalance, or other distributional properties. These are called dataset \meta. A distinct class of such estimators is the set of cheap to evaluate non-candidate algorithms called landmarks e.g. linear regression, logistic regression, or SVM. Their performance, or lack thereof, depends on their respective inductive biases and thereby can elucidate some properties of the dataset's topology. The indicative quality of these landmarks w.r.t. the performances of candidate algorithms, highly depends on the alignment of the geometries that the landmarks and candidate algorithms search for.

Complementary to \as, methods in \mfo \citep{kandasamy-nips16a, li-iclr17a, falkner-icml18a} use different types of fidelities such as dataset subset-sizes or number of epochs, summarized in the unrolling of algorithm's \lc, in order to cheaply evaluate candidate algorithms at various granularities. These fidelities are used to cheaply probe algorithms and draw conclusions about their final performances. There are, however, two major hurdles when utilizing fidelity information: 
\begin{inparaenum}[(i)]
    \item Acting on a fixed fidelity level rather than a learning curve (E.g. Successive Halving, BOHB \cite{falkner-icml18a}) requires that the relative performances need to be sufficiently indicative on that level. This pronounces the question of the reliability of the fidelity signal.
    \item A learning mechanism acting on multiple fidelities (as with partial learning curves) needs to be non-myopic to avoid fallacies such as a bias towards early performing algorithms or it might not recover.
\end{inparaenum}
While \metal \citep{vanschoren-arxiv18a} can alleviate myopia, it still requires the formulation of a meta-model and a mechanism to continually update it based on incoming evidence. Approaches that are less explicit in the use of their meta-prior form strong assumptions regarding the functional form of the \lc,  such as concavity or that the learning curve unfolds very similar in shape to previously observed ones \citep{leite-icml05}. These assumptions are difficult to justify and require theoretical as well as experimental rigor \citep{mohr-arxiv22}.

Exploiting \mfo, we enrich the set of landmarking estimators with landmarks derived from the candidate algorithms directly, thus, foregoing the need to rely on unreliable non-candidate algorithms. We, thereby, exchange the aforementioned alignment for a precision problem. Our \metal method, called Implicit Multi-Fidelity Algorithm Selection (\imfas), conditions on a vague representation of the dataset topology based on pre-computed \meta. It then refines this characterization using partially unrolled \lc of the algorithms. Finally, it uses the observed, but implicitly learned, relations between the inductive biases of the algorithms in the form of their \lc in the meta-training set to propagate this characterization into an expected ranking by means of this meta-prior. This alleviates myopia non-parametrically.

In summary, our contributions are the following: 
\begin{inparaenum}[(i)]
    \item We pose the \as problem as a \mf meta-learning few-shot ranking problem, which allows us to cheaply gather directly applicable and useful information about the new datasets' topology, bypassing former vicarious landmarking features.
    \item We use LSTMs \citep{hochreiter-nc97} and a differentiable ranking mechanism \citep{blondel2020fast}, to characterize a dataset's topology jointly by pre-computed \meta, the candidate algorithms' low fidelity approximations, and the meta experience of their relative progressions -- thus jointly and non-parametrically learning the \lc of algorithms on unseen datasets. 
\end{inparaenum}

\section{Background and Related Work}

Classical \as approaches learn mappings between datasets and algorithms w.r.t. performance based on pre-computed \meta, but these fail to sufficiently characterize the datasets and algorithms. Since landmarking approaches have shown considerable performance improvement \citep{pfahringer-icml00}, using a combination of pre-computed \meta and landmarks is conceptually appealing. Landmarks are not always superior since it is non-trivial to relate the landmarks' inductive biases to those of the candidates \citep{pfahringer-icml00}. Our approach sidesteps this by landmarking datasets directly based on the candidates algorithms' inductive biases.

The idea of exploiting partial learning curves has previously been explored in other settings. SAM~\citep{leite-icml05} non-parametrically matched partial learning curves using k-NN to the closest in terms of shape from the observed meta-learning curves, while \cite{rijn-ida15} reduced the cost of cross-validation by exploiting the similarity of the partially-observed rankings with those of the meta-datasets and using the most similar \lc as surrogates. \cite{mohr-arxiv21} extended this by terminating less promising candidates early on due to the abundance/redundancy of data by exploiting semi-parametric learning curves, under the assumption of concavity. They remain unbiased by not modeling meta-knowledge. \cite{leite-ecai10a} used a graph-based approach on a handcrafted performance distance metric between datasets and fidelities to account for ordinal rankings that originate from their repeated pairwise comparisons of algorithms across fidelities. In essence, they pursue a cheaper and meta-informed alternative to cross-validation that 
evaluates the collection of fidelities using SAM, and then actively schedules them in a cost-aware manner.

Two approaches orthogonal to ours, although similar in terms of setup, are Meta-REVEAL \citep{nguyen-ial21} and MetaBu \citep{rakotoarison-iclr22}. Meta-REVEAL focuses on scheduling the algorithms and fidelities through a Reinforcement Learning perspective by modeling it as a REVEAL game. The agent acts on a discrete action space that does not account for the correlations between learning curves. Consequently, using the estimates produced by \imfas can provide a pre-processed action space for their problem. MetaBu extends the classical \as idea by relating dataset \meta and algorithms' hyperparameters using a learned optimal transport map. They, however, ignore that the dataset meta-features cannot sufficiently characterize the dataset, thus, subsequently falling victim to the same fallacies as the methods mentioned before.

\section{Method}

We follow the \as set up with a portfolio $\mathcal{A}$ of algorithms from which one or a group of algorithms is to be recommended for a given dataset. We aim to learn a meta-model learning $\mathcal{D} \mapsto r(f_n)$ that enriches the dataset \meta $\mathcal{D}$ implicitly with landmarks of the collection of partially-observed algorithm performances $\{f_k\}_{k=1,\dots,g}$ where $g \ll n$ and can predict the rank of the final performances $f_n$ using the function $r$.

\imfas landmarks datasets using the collective of candidate algorithm's inductive biases. This is inspired by work in multi-dependent-label classification for \as \citep{wang-acm14, khan-ieee20}, which explicitly models the co-occurrence of algorithms (labels), conditioned on the dataset's \meta,  hinting towards patterns in the algorithms' inductive biases. \imfas tries to exploit this correlation as it unfolds along the fidelities. We combine information from the pre-computed \meta and a sequence of fidelity information using LSTMs \citep{hochreiter-nc97}. 

\begin{figure}[tb]
    \centering
    \includegraphics[width=0.9\textwidth]{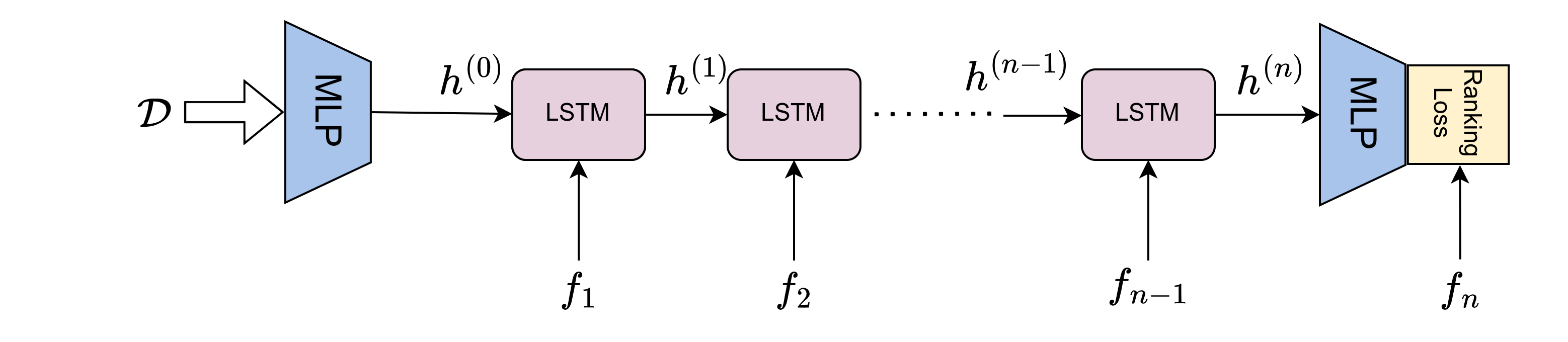}
    \caption{An overview of the method, initializing the LSTM using the encoded \meta and contextualizing it using the algorithm's performance vectors at the respective fidelities.}
    \label{fig:lstm_method}
\end{figure}

Since LSTMs fundamentally combine memory-based representations with contextual information fed at every step of the sequence, we encode the dataset \meta to initialize the hidden state. This is done by feeding the dataset \meta $\mathcal{D}$ to an encoder Multi-Layer Perceptron (MLP). Given $n$ fidelities going from the cheapest to full training of a model, we forward propagate the LSTM $n-1$ times. The output of the LSTM is again encoded to a vector of $|\mathcal{A}|$ dimensions using an MLP. The final output values are differentiably sorted and ranked \citep{blondel2020fast}. Finally, the ranks are compared against a ranking of \ground performances in the final fidelity as depicted in Figure~\ref{fig:lstm_method}. 

To create a loss for optimization, we calculate the spearman correlation \citep{zar-biostats05} between the predicted ranking $r_p$ and the \ground ranking $r_g$ and then complement this value to get the loss $ \mathcal{L}= 1 - \frac{Cov(r_p, r_g)}{\sigma_{r_p}\sigma_{r_g}}$, where $\sigma_{r_p}$ and $\sigma_{r_g}$ are the respective standard deviations of the rank variables. Minimizing this quantity amounts to maximizing the positive correlation between the predicted ranking and the \ground. The gradients of the differentiable ranking function are then propagated backward. During meta-test time, we repeat the same process but with the crucial difference that instead of unrolling the LSTM for all fidelities, we only unroll it up to a fraction of the fidelity level. This helps us examine whether the trained meta-model can approximate the full-fidelity rank with partial fidelity information, indicated by a low value of the test loss. Please refer to Appendix~\ref{app:impl} for implementation details

Intuitively, the hidden dimensions $d$ of LSTM are constant, the hidden representation at each step can be thought of as a point that moves in the latent space $h^{(i)} \in   \mathbb{R}^d$. Thus, given an encoding of the dataset \meta in that latent space as initialization, unrolling the LSTM each time based on the current $f_i$ is akin to tracking a trajectory of this point. During each unroll step, the LSTM acts like a transition operator on that space, progressively conditioning on the obtained fidelity information. Applying the final decoder MLP on any of the $h^{(i)}$ amounts to the LSTM's current expectation $\hat{f}_n$ conditioned on $\{f_j\}_{j=1,\dots,i}$. This operator's learned weights encapsulate all the meta-knowledge about how the algorithms' trajectories correlate conditioned on the dataset's \meta. It acts as meta-prior, similar to \cite{ravi-iclr16}, during test time which combats myopia in an exclusively data-driven and non-parametric manner. We merely need to apply this operator to condition our expectation towards $\hat{f}_n$ on the partially available fidelity information.

\section{Experiments}

\paragraph{Benchmarks and Baseline} We evaluate our approach\footnote{The implementation of \imfas can be found on GitHub \url{https://github.com/automl/IMFAS}} on the YAHPO-Gym surrogate benchmark \citep{pfisterer-automlconf22}, specifically the rbv2 algorithm instances, which is a collection of separate \hpo surrogate benchmarks, each for a specific algorithm. All of the rbv2 instances consider dataset subset size as fidelity type. Crucially, we model different algorithms as separate instantiations of the hyperparameters of an algorithm, which has been shown to sufficiently alter inductive bias to be relevant to our problem \citep{li-nips21}. We additionally test \imfas on LCBench \citep{zimmer-ieee2021a}, an epoch-based learning curve benchmark that provides extensive meta-training data for different MLP architectures and hyperparameters. Finally, to validate whether the learned rankings are sensible, we employ a meta-ignorant, non-parametric, and myopic successive halving \citep{li-iclr17a} as a baseline. It is altered to produce an absolute ranking in the following fashion: The level of termination (fidelity) generates a tied ordinal ranking, where each level's ties are broken by considering the observed performances at that level. As a result, we obtain an absolute ranking, respecting the amount of available information on the levels. This predicted ranking is compared against the \ground.

\paragraph{Setup} We evaluated the agent on the aforementioned benchmarks by training it for $300$ epochs on the full fidelity information. The evaluations were performed by holding out $20\%$ of the data for testing. Additionally, the \sh baseline has been computed on the test data that was held out. We report the results as the average and standard deviation across $5$ seeds. 

\paragraph{Results} We summarize our results in Table~\ref{tab:results}. The metric used for the rbv2 suite is the F1 score and the one for LCBench is the validation accuracy. Notice that the Spearman correlation generally increases with an increasing amount of fidelity information. However, the differences between the final correlations produced at 50\% fidelities and at 100\% fidelities during test time are generally lower, indicating that the transfer helps the model in approximating a correlation close to the \ground using cheaper approximations. \imfas is able to beat \sh in most cases (highlighted in bold) at fidelities below 100\%. 


\begin{table}[htb]
\centering
\caption{Summary of resulting correlations during test time given the amount of available fidelity information in (\%) of the fidelity sequence against \sh. In bold are the first fidelities at which the baseline is surpassed.}
\label{tab:results}
\addtolength{\tabcolsep}{-3pt}    
\begin{tabular}{l|ccc|c}
\toprule
\textbf{Dataset} & \textbf{10\%}     & \textbf{20\%}     & \textbf{50\%}    & \textbf{SH}\\
\midrule
rbv2\_super      & $0.606 \pm 0.021$ & $\mathbf{0.698 \pm 0.070}$ & $0.759 \pm 0.021$ & $0.694 \pm 0.073$\\
rbv2\_svm        & $0.431 \pm 0.013$ & $0.504 \pm 0.182$ & $\mathbf{0.660 \pm 0.104}$ & $0.641 \pm 0.079$\\
rbv2\_xgboost    & $0.473 \pm 0.054$ & $0.596 \pm 0.034$ & $\mathbf{0.769 \pm 0.010}$ & $0.668 \pm 0.096$\\
rbv2\_ranger     & $0.454 \pm 0.028$ & $0.454 \pm 0.144$ & $\mathbf{0.552 \pm 0.092}$ & $0.500 \pm 0.084$\\
rbv2\_rpart      & $0.391 \pm 0.342$ & $\mathbf{0.554 \pm 0.237}$ & $0.749 \pm 0.086$ & $0.351 \pm 0.093$\\
rbv2\_aknn       & $\mathbf{0.539 \pm 0.209}$ & $0.633 \pm 0.059$ & $0.704 \pm 0.057$ & $0.497 \pm 0.068$\\
\hline
lcbench          & $0.616 \pm 0.045$ & $0.710 \pm 0.041$ & $\mathbf{0.759 \pm 0.031}$ & $0.641 \pm 0.079$\\
\bottomrule
\end{tabular}
\end{table}

\begin{figure}[htb]
    \centering
    \includegraphics[width=0.7\textwidth]{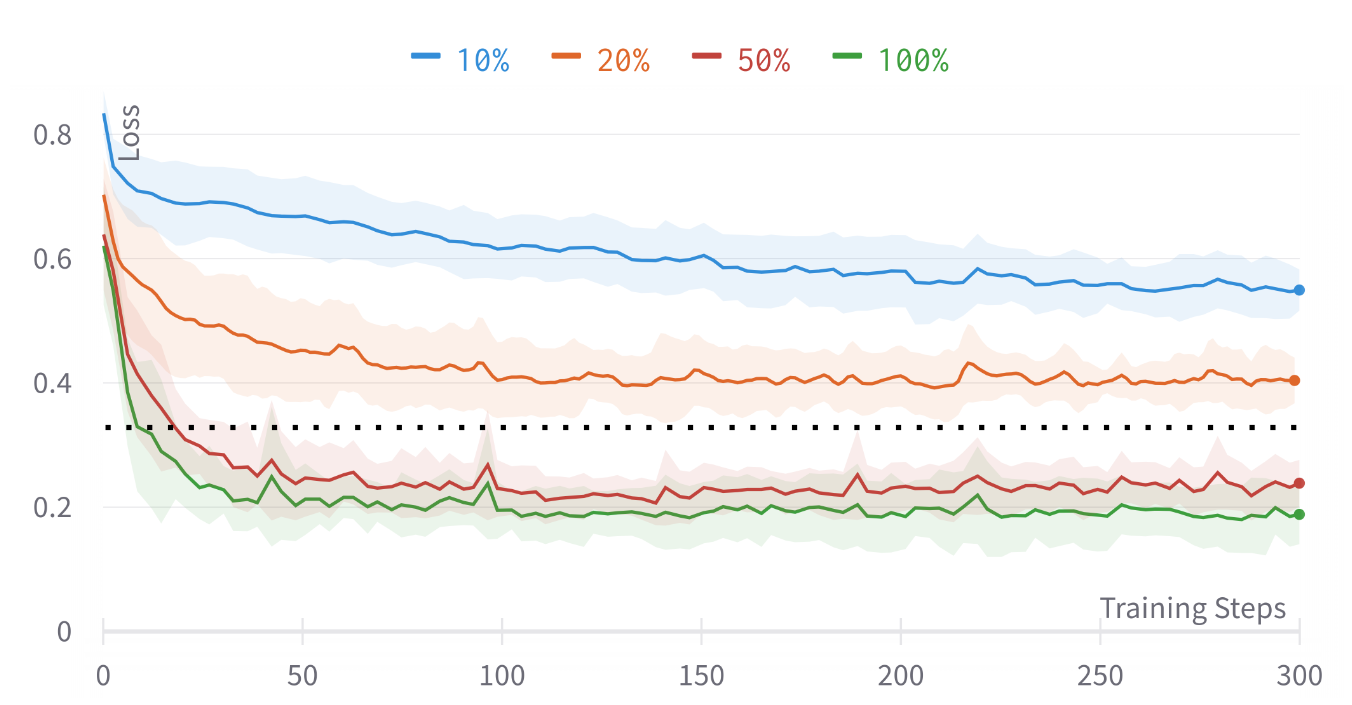}
    \caption{The loss progression in terms of a learning curve for \texttt{rbv2\_xgboost}. By only using $50\%$ of the fidelity sequence, our model approximates the the final ranking possible with the full sequence. The dotted line shows the value produced by \sh for this dataset}
    \label{fig:rbv2_svm}
\end{figure}

Figure~\ref{fig:rbv2_svm} shows an exemplary plot for  rbv2\_xgboost dataset and we see that the model is able to approximate the performance of the 100\% fidelities by only unrolling for 50\%, as compared to when it is fed only 10\% or 20\% of the sequence during test time. This shows positive transfer, since the higher fidelities can be approximated from the cheaper. 

\section{Conclusion and Future work}

We present a unifying outlook towards \as through a \mf lens and present a method to capture the fidelity information implicitly and non-parametrically using LSTMs with a differentiable ranking loss. We test this approach on diverse benchmarks and show that lower fidelities captured in this manner are often sufficient to implicitly produce a ranking comparable to the \ground. We finally present an analysis of how the sufficient fidelities depend on the benchmark being tested. Our approach, however, has certain limitations which we plan to address in future work: 
\begin{inparaenum}[(i)]
\item The current methodology does not completely support hyperparameter settings, albeit their features, which vicariously describe the inductive biases, are captured implicitly by the learning curves. Exploiting this information allows extending \imfas to \hpo across multiple algorithms, thus addressing the full CASH problem \citep{thornton-kdd13a}.
\item The requirement of a full grid of algorithms and datasets in the meta-dataset can be reduced using an interpolation strategy based on partially overlapping dataset-algorithm combinations, thus, improving scalability.
\item Estimates regarding uncertainty resulting from the fidelity spacing and varying degree of contained information can be propagated to potentially better inform the fidelity roll-out. 
\end{inparaenum}

\section*{Acknowledgement}
The authors acknowledge financial support by the Federal Ministry for Economic Affairs and Energy of Germany in the project CoyPu under Grant No. 01MK21007L.

\bibliography{bib/local,bib/lib,bib/proc,bib/strings}

\newpage

\appendix
\section{Implementation Details}
\label{app:impl}

\paragraph{LSTM Details} The general hyperparameter configuration of the LSTM has been shown in Table~\ref{tab:lstmgen}. The LSTM was implemented using the pytorch\citep{pytorch-neurips19a} library, and the differentiable ranking losses have been implemented using torchsort \citep{blondel2020fast}. The training was performed on a 16-core CPU. The parameters that changed within tests were the \texttt{batch size} and \texttt{learning rate}, which have been reported in Table~\ref{tab:lstmdata}. The hyperparameters were tuned by elucidating the configuration space using Hyperband \citep{li-iclr17a} and then checking the ones that work from the reduced set of configurations.

\begin{table}[htb]
\centering
\caption{Table showing the general hyperparameter configuration of lstm}
\label{tab:lstmgen}
\addtolength{\tabcolsep}{-3pt}    
\begin{tabular}{l|c}
\hline
\textbf{Hyperparameter} & \textbf{Value}\\
\hline
Number of Layers    & $2$ \\
Hidden dimensions of Encoder & $(300, 200)$ \\
Hidden dimensions of readout & $(200, |\mathcal{A}|)$ \\
\hline
\end{tabular}
\end{table}

\begin{table}[htb]
\centering
\caption{Table showing the general hyperparameter configuration of LSTM}
\label{tab:lstmdata}
\addtolength{\tabcolsep}{-3pt}    
\begin{tabular}{l|cc}
\toprule
\textbf{Dataset} & \textbf{Batch Size} & \textbf{Learning Rate}\\
\midrule
rbv2\_super      &$10$ &$0.001$\\
rbv2\_svm        &$10$ &$0.0005$ \\
rbv2\_xgboost    &$10$ &$0.0005$ \\
rbv2\_ranger     &$10$ &$0.001$  \\
rbv2\_rpart      &$10$ &$0.001$  \\
rbv2\_aknn       &$10$ &$0.0005$ \\
lcbench         &$8$   &$0.001$   \\
\bottomrule
\end{tabular}
\end{table}

\paragraph{YAHPO Benchmark} Yahpo consists of multiple \hpo benchmarks of which we chose those named in the form rbv2\_$<$algorithm$>$ including $<$algorithm$>$: svm, xgboost, aknn, ranger, rpart, super. Noteworthy, 'super' is a full ML pipeline, joining the search spaces of rpart, glmnet, ranger, xgboost hierarchically, thereby encapsulating a CASH problem. Since Yahpo is a surrogate benchmark, we sample 50 hyperparameter configurations for each of the rbv2 instances respectively using a Latin Hypercube Design. These configurations are held constant for each instance across all its datasets. With the exception of 'super', sampling a hyperparameter space in this way likely will introduce correlating learning curves. The 'super' instance combines the latter with heterogeneous algorithms, likely differing in their inductive biases.

\paragraph{LCBench Benchmark} Using the raw LCBench, which reports its real evaluations, we sample the table of hyperparameters in a manner, that produces a set of diverse and known to be performant configurations. We do this by considering each dataset's top-k performing algorithms and building the union over those resulting configurations. Using $k=3$, we arrive at 58 candidate algorithms.

\end{document}